\pdfoutput=1

\documentclass[11pt]{article}

\usepackage{acl}

\usepackage{times}
\usepackage{latexsym}

\usepackage[T1]{fontenc}

\usepackage[utf8]{inputenc}
\usepackage{graphicx}

\usepackage{microtype}
\usepackage{booktabs, multirow} 
\usepackage{amsfonts, amsmath, amssymb}
\usepackage{xurl}
\usepackage{soul}

\title{HiJoNLP at SemEval-2022 Task 2: Detecting Idiomaticity of Multiword Expressions using Multilingual Pretrained Language Models}

\author{
Minghuan Tan\\
School of Computing and Information Systems \\
Singapore Management University\\
mhtan.2017@phdcs.smu.edu.sg
}

\begin{document}
\maketitle
\begin{abstract}
This paper describes an approach to detect idiomaticity only from the contextualized representation of a MWE over multilingual pretrained language models.
Our experiments find that larger models are usually more effective in idiomaticity detection. 
However, using a higher layer of the model may not guarantee a better performance.
In multilingual scenarios, the convergence of different languages are not consistent and rich-resource languages have big advantages over other languages.
\end{abstract}

\section{Introduction}

In the past several years, there have been breakthroughs in a variety of natural language processing tasks with the power of pretrained language models.
These include but are not limit to question answering~\cite{devlin-etal-2019-bert}, language generation~\cite{radford2018improving,radford2019language} and machine translation~\cite{liu-etal-2020-multilingual-denoising}.
However, it's still not clear whether pretrained language models have the ability in capturing the meanings of multiword expressions (MWEs), especially idioms.
Given the prevalent usage of idioms in different languages, identifying the correct meaning of a phrase in a certain context is crucial for many downstream tasks including sentiment analysis~\cite{WILLIAMS20157375}, automatic spelling correction~\cite{horbach-etal-2016-corpus} and machine translation~\cite{isabelle-etal-2017-challenge}.

In literature, idiomaticity detection has been a research topic drawing much attention from the NLP community.
MWEs which have both an idiomatic interpretation and a literal interpretation are also referred as Potentially Idiomatic Expressions~(PIEs), for example, \emph{spill the beans}.
There has been both supervised~\cite{sporleder-li-2009-unsupervised} and unsupervised~\cite{haagsma-etal-2018-side,kurfali-ostling-2020-disambiguation} approaches to solve this problem.
For example, \newcite{feldman-etal-cicling-2013-automatic} treated idiom recognition as outlier detection, which does not rely on costly annotated training data.
\newcite{peng-etal-2014-classifying} incorporated the affective hypothesis of idioms to facilitate the identification of idiomatic operations. 

Due to the limited understanding of how pretrained language models may handle representation of phrases, a series of works are proposed to investigate phrase composition from their contextualized representations.
\citet{yu-ettinger-2020-assessing} conduct analysis of phrasal representations in state-of-the-art pre-trained transformers and find that phrase representation in these models still relies heavily on word content, showing little evidence of nuanced composition. 
\citet{shwartz-dagan-2019-still} confirm that contextualized word representations perform better than static word embeddings, more so on detecting meaning shift than in recovering implicit information.
Therefore, it remains a challenging problem to resolve the idiomaticity of phrases.

Specifically on idiomaticity, recent approaches are trying to further diagnose pretrained language models using new metrics and datasets.
\citet{garcia-etal-2021-assessing} analyse different levels of contextualisation to check to what extent models are able to detect idiomaticity at type and token level.
\citet{garcia-etal-2021-probing} propose probing measures to assess Noun Compound~(NC) idiomaticity and conclude that idiomaticity is not yet accurately represented by contextualised models.
AStitchInLanguageModels~\cite{tayyar-madabushi-etal-2021-astitchinlanguagemodels-dataset} design two tasks to first test a language model’s ability to detect idiom usage, and the effectiveness of a language model in generating representations of sentences containing idioms.
\citet{tan-jiang-2021-bert} conduct two probing tasks, PIE usage classification and idiom paraphrase identification, suggesting that BERT indeed is able to separate the literal and idiomatic usages of a PIE with high accuracy and is also able to encode the idiomatic meaning of a PIE to some extent.
However, there's still much more to explore in idiomaticity.

Based upon AStitchInLanguageModels~\cite{tayyar-madabushi-etal-2021-astitchinlanguagemodels-dataset}, SemEval-2022 Task2~\cite{tayyarmadabushi-etal-2022-semeval} is proposed with a focus on multilingual idiomaticity.
The task is arranged consisting the two subtasks:
\begin{enumerate}
    \item Subtask A: A binary classification task aimed at determining whether a sentence contains an idiomatic expression.
    \item Subtask B: Pretrain or finetune a model which is expected to output the correct Semantic Text Similarity (STS) scores between sentence pairs, whether or not either sentence contains an idiomatic expression.
\end{enumerate}

In this paper, we focus on Subtask A and investigate how the span representation of a MWE can tell about its idiomaticity.
We extend one of the monolingual idiomaticity probing method~\cite{tan-jiang-2021-bert} to multilingual scenario and compare multiple settings using  multi-lingual BERT~(mBERT)~\cite{devlin-etal-2019-bert} and XLM-R~\cite{conneau-etal-2020-unsupervised}.
Following \citet{yu-ettinger-2020-assessing}, we also consider variations of phrase representations across models, layers, and representation types.
Different from them, we use more representation types to conduct the experiments.

Our main conclusion from these experiments are two folds: 
\begin{enumerate}
    \item Larger models are usually more effective in idiomaticity detection. However, a higher layer may not contribute more to the idiomaticity detection task, or more contextualization does not guarantee a better performance.
    \item For multilingual scenario, the convergence of different languages are not consistent. Rich resource languages have initiative advantages over other languages.
\end{enumerate}

\section{System Overview}

\subsection{Subtask A}

For Subtask A, to test models' ability to generalise, both zero-shot and one-shot settings are considered.
\begin{enumerate}
    \item zero-shot: PIEs in the training set are completely disjoint from those in the test and development sets.
    \item one-shot: one positive and one negative training examples for each MWE in the test and development sets
\end{enumerate}
Note that the actual examples in the training data are different from those in the test and development sets in both settings. 

\paragraph{Data}


Each row of the data of Subtask A has attributes like language and the potentially idiomatic MWE. 
The "Target" is the sentence that contains this MWE. 
The previous and next sentences for context are also provided. 
The label provides the annotation of that row, and a label of 0 indicates "Idiomatic" and a label of 1 indicates "non-idiomatic", including proper nouns.

\paragraph{Baseline} The baseline model~\cite{tayyarmadabushi-etal-2022-semeval} is based on mBERT.
In the zero-shot setting, the model uses the context (the sentences preceding and succeeding the one containing the idioms) and does not add the idiom as an additional feature (in the “second input sentence”). 
In the one shot setting, the model is trained on both the zero-shot and one-shot data, but exclude the context (the sentences preceding and succeeding the one containing the idioms) and add the idiom as an additional feature in the “second sentence”.

\begin{figure*}[!ht]
\centering
\includegraphics[width=0.8\linewidth]{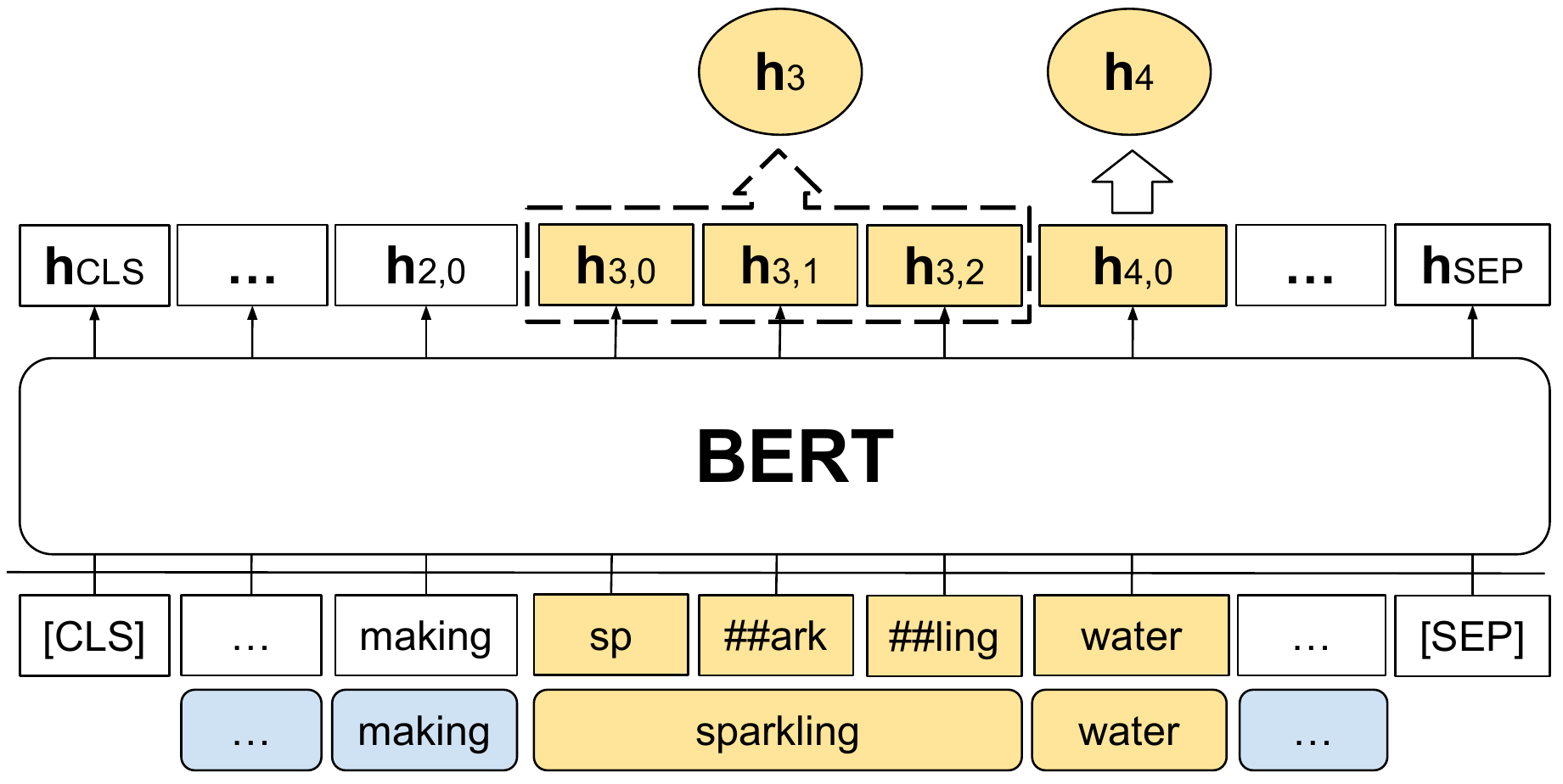}
\caption{Mismatched transformer-based span representation.} 
\label{fig:span}
\end{figure*}

\subsection{Span-based Model}
While the common practice for classification tasks using pretrained language models usually needs concatenation of text sequences, this does not tell us enough information how representations of MWEs may lead to the change of performance.
Therefore, in this work, we focus on the contextualized representations of MWEs to predict its idiomaticity.

\paragraph{Problem Formulation} Consisting with the definition in~\cite{tan-jiang-2021-bert} , given a sentence denoted as $(w_1, w_2, \dots, w_n)$, which contains a MWE with $m$ words denoted as $(w_i, \dots, w_{i+m-1})$,
The task is to decide whether the MWE is used with its \textit{literal} meaning or its \textit{idiomatic} meaning, or if a sentence contains an idiomatic expression as describe in the task.

\paragraph{Span Identification}
In this work, our method requires a pair of span indices of the target MWE to extract their hidden representation from the encoded sequence.
However, in this task, no such indices is offered explicitly from the dataset.
We empirically find these indices by using editing distances in characters between the MWE and the sentence.
This method works for most of the cases.

\paragraph{Span Representation} 
For each MWE, we have a pair of span offsets in the original context.
We use an $L$-layer BERT to process the tokenized context by prepending $\texttt{[CLS]}$ to the beginning and appending $\texttt{[SEP]}$ to the end.
Let $\mathbf{h}^k_i \in \mathbb{R}^d$ denote the hidden vector produced by the $k$\-th layer of BERT representing $w_i$.
We extract the hidden representations of the span to get its contextualized representations.
For each MWE, we get a sequence of hidden vectors at the $k$-th layer for the $m$ tokens inside this MWE as follows: $\mathbf{p}^k = (\mathbf{h}^k_i, \mathbf{h}^k_{i+1}, \ldots, \mathbf{h}^k_{i+m-1})$.

In transformer-based models, a word might be tokenized into several pieces.
We adopt the mismatched tokenization trick offered by Allennlp~\footnote{\url{https://github.com/allenai/allennlp}} to reconstruct its hidden vector.
The hidden vector will be the average embeddings of constituent pieces.
The mismatched encoding is illustrated in Figure~\ref{fig:span}.

We represent the target MWE using the span by six different kinds of combinations of the span's words.
The first four of them are only using their endpoints.
We use $\mathbf{x}=\mathbf{h}^k_i$ to denote the start of the span and $\mathbf{y}=\mathbf{h}^k_{i+m-1}$ to denote the end of the span.
\begin{enumerate}
    \item \textbf{x,y} The span is represented by a direct concatenation of two endpoints.
    \item \textbf{x,y,x-y} The span is represented by a direct concatenation of two endpoints and the difference of them.
    \item \textbf{x,y,x*y} The span is represented by a direct concatenation of two endpoints and the elementwise product of them.
    \item \textbf{x,y,x*y,x-y} The span is represented by a direct concatenation of two endpoints, the elementwise product and the difference of them.
    \item \textbf{SelfAttentive}  We firstly compute an unnormalized attention score for each word in the document.
    Then we compute spans representations with respect to these scores by normalising the attention scores for words inside the span.
    \item \textbf{MaxPooling} A span is represented through a dimension-wise max-pooling operation.
    Given a span, the resulting value of a dimension is using the maximum value of this dimension across all the span tokens.
\end{enumerate}

\paragraph{Span Classification} We use a binary linear classifier upon the span representation.




\section{Experiments}
In this paper, we want to test how the pretrained model, the transformer layer and the representation type, affect performance of idiomaticity detection.

\begin{table*}[!ht]
\centering
\begin{tabular}{lcrrrrrrr}\toprule
Model & Type  &Layer & &EN &PT &GL &Avg \\\midrule
mBERT~\cite{tayyarmadabushi-etal-2022-semeval} &- &12 & &70.70 &68.03 &50.65 &65.40 \\\midrule
mBERT &x,y,x-y &12 & &76.24 &72.27 &64.27 &72.85 \\
XLM-R &x,y &8 & &\textbf{77.62} &71.61 &64.88 &72.68 \\
XLM-R-L &x,y,x-y &24 & &75.22 &\textbf{75.80} &\textbf{69.01} &\textbf{74.66} \\
\bottomrule
\end{tabular}
\caption{Experiment results of zero-shot setting for different multilingual pretrained models, in macro F1 score. }
\label{tab:zero-shot-results}
\end{table*}

\begin{table*}[!ht]
\centering
\begin{tabular}{lcrrrrrrr}\toprule
Model &Type  &Layer & &EN &PT &GL &Avg \\\midrule
mBERT~\cite{tayyarmadabushi-etal-2022-semeval} &- &12 & &88.62 &86.37 &81.62 &86.46 \\\midrule
mBERT &MaxPooling &8 & &86.59 &85.82 &85.77 &86.63 \\
XLM-R &MaxPooling &8 & &89.49 &83.71 &82.19 &86.17 \\
XLM-R-L &x,y,x*y,x-y &24 & &\textbf{91.26} &\textbf{86.96} &\textbf{89.06} &\textbf{89.79} \\
\bottomrule
\end{tabular}
\caption{Experiment results of one-shot setting for different multilingual pretrained models, in macro F1 score. }
\label{tab:one-shot-results}
\end{table*}

\subsection{Settings}
This subtask is evaluated using the Macro F1 score between the gold labels and model predictions (see the details in the evaluation script). 

All the multilingual pretrained langauge models are hold by Huggingface, including mBERT\footnote{BERT multilingual base (cased)
: \url{https://huggingface.co/bert-base-multilingual-cased}}, XLM-R\footnote{XLM-RoBERTa (base-sized model): \url{https://huggingface.co/xlm-roberta-base}} and XLM-R-L\footnote{XLM-RoBERTa (large-sized model)
: \url{https://huggingface.co/xlm-roberta-large}}.

Since we are focusing on comparison of span representation across different layers and representation types, we conduct experiments with the 4-th, 8-th and 12-th layer of mBERT and XLM-R and the 8-th, 12-th and 24-th layer of XLM-R-L.
All six representation types are considered for each layer-based models.

We run most of our experiments with an NVIDIA 1080ti GPU with 11GB memory, and use a NVIDIA A100 for XLM-R-L-based experiments.
We finetune each experiment for 10 epochs with the learning rate set to 5e-5.
We notice that the training process converges with training accuracy 1 in a short period.
To reduce the effect of overfitting, we use a dropout probability of 0.5 before the classification layer.
Our code is built over Allennlp2 and will be released on Github\footnote{\url{https://github.com/VisualJoyce/CiYi}}.

\subsection{Results and Analyses for Subtask A}

We list the overall experiment results in Table~\ref{tab:all-results} in the Appendix.
The table contains three main parts with each part showing the detailed experiment results for a multilingual pretrained language model.
In each part, we test all six combinations of span representations using encoded sequences from different layers.
To better illustrate our major conclusions, we select the best settings for each multilingual model from Table~\ref{tab:all-results}, and rearrange the zero-shot results to Table~\ref{tab:zero-shot-results} and one-shot results to  Table~\ref{tab:one-shot-results}.

Table~\ref{tab:zero-shot-results} shows us that using only endpoints of the span can be effective in predicting its idiomaticity and representation type \textbf{x,y,x-y} is a good choice for the zero-shot setting.
We think representation using only endpoints is working well might due to most of the MWEs in current dataset consist of two words.

Table~\ref{tab:one-shot-results} shows us that representation type \textbf{MaxPooling} is a good choice for the one-shot setting and the best performance may be achieved using middle layers.

Combining both zero-shot setting and one-shot setting, we find that larger models are usually more effective in idiomaticity detection.
For a specific pretrained model, using contextualized representation from a higher layer may not guarantee a better performance.
For example, from the perspective of overall score for the One Shot scenario, the highest scores are all reached at the 8-th layer.
However, we didn't observe a consistent advantage of using a specific representation type across different models and layers.

From the perspective of language, span-based models are achieving relative larger gains in both settings for GL.
On one hand, the corpus used for training pretrained language models is not balanced across different languages.
For example, in XLM-R, data from EN is several times than that of PT and hundrands times than that of GL.
The data for GL may just surpass a minimal size for learning a BERT model and restricts performance in both settings for GL compared with PT and EN.
On the other hand, this tells us that better span representation still help in detection of idiomaticity.


\subsection{Endpoints-based Representation}
This work focuses on the contextualized representation of the span of a target MWE.
As pointed out by others, phrase representations, especially idioms, are not always compositional and rely more than the constituent words in the span.
Not to mention, it is a much easier case which only uses the endpoints of the span.
However, in both zero-shot setting and one-shot setting, we notice that endpoints-based methods works almost as well.
We suspect this may due to the following reasons:
(1) Endpoints of MWEs are highly correlated with these MWEs and can be very indicative about their representation.
(2) Most of the MWEs covered in this dataset contain two words.

\section{Conclusion}
In conclusion, our experiments find that larger models are usually more effective in idiomaticity detection.
And for a specific pretrained model, using contetualized representation from a higher layer may not guarantee a better performance.
As the data used for multilingual pretrained language models is not well-balanced, rich resource languages have significant advantages over other languages.
In the future, with the community contributing stronger language models with more balanced language distribution and more multilingual idiom-annotated datasets, idiomaticity detection still has large potentials to be explored from more angles.


\bibliography{anthology,custom}
\bibliographystyle{acl_natbib}

\appendix


\begin{table*}[t]\centering
\small
\begin{tabular}{lrrrrrrrrrrrrr}\toprule
\multirow{3}{*}{Model} &\multirow{3}{*}{Type} &\multirow{3}{*}{Layer} & &\multicolumn{4}{c}{Zero Shot} & &\multicolumn{4}{c}{One Shot} \\\cmidrule{5-8}\cmidrule{10-13}
& & & &EN &PT &GL &Avg & &EN &PT &GL &Avg \\\midrule
mBERT &- &12 & &70.70 &68.03 &50.65 &65.40 & &88.62 &86.37 &81.62 &86.46 \\
\midrule
mBERT &x,y &4 & &75.11 &69.63 &64.20 &72.49 & &86.32 &85.17 &76.50 &83.84 \\
mBERT &x,y,x-y &4 & &73.69 &71.69 &57.96 &70.31 & &86.51 &85.68 &77.04 &84.25 \\
mBERT &x,y,x*y &4 & &76.76 &70.67 &60.27 &71.69 & &87.76 &\ul{86.15} &80.16 &85.93 \\
mBERT &x,y,x*y,x-y &4 & &75.54 &\ul{73.56} &60.18 &71.62 & &89.28 &85.16 &80.21 &86.17 \\
mBERT &SelfAttentive &4 & &72.13 &73.19 &62.16 &70.79 & &85.48 &82.86 &78.76 &83.50 \\
mBERT &MaxPooling &4 & &71.27 &73.11 &58.46 &69.49 & &85.23 &83.40 &76.56 &82.93 \\
& & & & & & & & & & & & \\
mBERT &x,y &8 & &75.95 &68.49 &\ul{65.07} &72.21 & &85.87 &84.91 &81.21 &84.97 \\
mBERT &x,y,x-y &8 & &72.45 &66.88 &61.95 &69.11 & &86.86 &83.95 &82.47 &85.41 \\
mBERT &x,y,x*y &8 & &75.14 &67.73 &61.81 &70.49 & &86.55 &81.82 &81.29 &84.23 \\
mBERT &x,y,x*y,x-y &8 & &72.59 &73.18 &61.91 &70.87 & &86.09 &84.21 &81.30 &84.79 \\
mBERT &SelfAttentive &8 & &73.84 &68.60 &62.22 &69.78 & &\ul{89.69} &82.72 &83.65 &86.46 \\
mBERT &MaxPooling &8 & &77.24 &68.59 &62.16 &71.89 & &86.59 &85.82 &\ul{85.77} &\ul{86.63} \\
& & & & & & & & & & & & \\
mBERT &x,y &12 & &76.31 &70.77 &58.80 &70.36 & &86.45 &84.06 &79.69 &84.47 \\
mBERT &x,y,x-y &12 & &76.24 &72.27 &64.27 &\ul{72.85} & &85.91 &85.19 &82.60 &85.47 \\
mBERT &x,y,x*y &12 & &74.04 &71.76 &64.24 &71.65 & &87.58 &84.27 &79.92 &85.00 \\
mBERT &x,y,x*y,x-y &12 & &\ul{78.63} &69.01 &62.91 &72.62 & &86.66 &85.24 &79.05 &84.75 \\
mBERT &SelfAttentive &12 & &75.03 &69.71 &60.75 &70.32 & &86.31 &82.62 &83.69 &85.14 \\
mBERT &MaxPooling &12 & &75.33 &71.08 &59.00 &69.90 & &88.37 &85.38 &81.19 &86.07 \\
\midrule
XLM-R &x,y &4 & &80.70 &65.29 &54.57 &68.82 & &89.26 &80.22 &73.15 &82.48 \\
XLM-R &x,y,x-y &4 & &79.49 &67.51 &54.98 &69.20 & &89.27 &82.26 &74.10 &83.47 \\
XLM-R &x,y,x*y &4 & &78.66 &70.77 &57.66 &71.19 & &88.38 &79.47 &70.03 &80.79 \\
XLM-R &x,y,x*y,x-y &4 & &74.10 &67.11 &56.48 &68.49 & &88.30 &81.13 &73.96 &82.73 \\
XLM-R &SelfAttentive &4 & &\textbf{81.51} &70.99 &55.49 &71.91 & &88.46 &80.84 &74.82 &82.78 \\
XLM-R &MaxPooling &4 & &78.57 &67.48 &59.38 &70.69 & &88.97 &81.83 &79.99 &84.81 \\
& & & & & & & & & & & & \\
XLM-R &x,y &8 & &77.62 &71.61 &64.88 &\ul{72.68} & &89.18 &80.98 &78.43 &84.22 \\
XLM-R &x,y,x-y &8 & &76.86 &66.31 &60.52 &69.38 & &86.57 &81.33 &72.91 &81.51 \\
XLM-R &x,y,x*y &8 & &73.51 &62.41 &55.22 &65.02 & &87.58 &81.31 &75.57 &82.73 \\
XLM-R &x,y,x*y,x-y &8 & &77.70 &70.43 &\ul{65.19} &72.43 & &87.74 &78.24 &80.44 &83.36 \\
XLM-R &SelfAttentive &8 & &78.43 &68.01 &61.40 &71.08 & &89.03 &83.19 &75.55 &83.82 \\
XLM-R &MaxPooling &8 & &76.61 &68.45 &64.50 &71.35 & &89.49 &\ul{83.71} &\ul{82.19} &\ul{86.17} \\
& & & & & & & & & & & & \\
XLM-R &x,y &12 & &76.95 &66.35 &57.73 &68.54 & &86.66 &81.73 &77.73 &83.15 \\
XLM-R &x,y,x-y &12 & &75.98 &63.26 &55.31 &66.31 & &86.12 &79.82 &73.51 &80.92 \\
XLM-R &x,y,x*y &12 & &77.70 &70.18 &61.05 &71.05 & &87.65 &79.54 &74.17 &81.83 \\
XLM-R &x,y,x*y,x-y &12 & &78.07 &71.51 &59.04 &70.91 & &88.19 &82.09 &76.30 &83.45 \\
XLM-R &SelfAttentive &12 & &76.16 &70.54 &62.92 &71.36 & &\ul{90.05} &79.77 &77.26 &83.82 \\
XLM-R &MaxPooling &12 & &74.98 &\ul{75.23} &63.80 &72.31 & &85.67 &81.03 &75.50 &81.88 \\
\midrule
XLM-R-L &x,y &8 & &79.10 &72.78 &61.68 &72.82 & &91.89 &85.56 &75.63 &85.86 \\
XLM-R-L &x,y,x-y &8 & &76.96 &59.09 &57.83 &68.44 & &89.08 &85.94 &76.44 &85.25 \\
XLM-R-L &x,y,x*y &8 & &73.51 &62.41 &55.22 &65.02 & &87.58 &81.31 &75.57 &82.73 \\
XLM-R-L &x,y,x*y,x-y &8 & &80.19 &71.09 &62.12 &73.45 & &91.92 &81.79 &72.77 &84.06 \\
XLM-R-L &SelfAttentive &8 & &77.25 &71.92 &59.94 &70.56 & &\textbf{92.66} &84.59 &77.57 &86.37 \\
XLM-R-L &MaxPooling &8 & &77.83 &70.92 &61.18 &71.40 & &89.85 &81.79 &69.14 &81.71 \\
& & & & & & & & & & & & \\
XLM-R-L &x,y &12 & &76.92 &70.40 &60.11 &70.17 & &90.26 &85.19 &82.76 &87.10 \\
XLM-R-L &x,y,x-y &12 & &77.48 &67.52 &60.25 &69.85 & &92.24 &81.30 &81.00 &86.30 \\
XLM-R-L &x,y,x*y &12 & &80.54 &65.49 &55.46 &68.72 & &91.43 &83.91 &78.78 &86.00 \\
XLM-R-L &x,y,x*y,x-y &12 & &79.77 &69.66 &60.84 &71.54 & &90.28 &84.42 &83.46 &87.14 \\
XLM-R-L &SelfAttentive &12 & &78.13 &74.44 &61.92 &72.63 & &90.48 &86.23 &78.90 &86.43 \\
XLM-R-L &MaxPooling &12 & &\ul{80.68} &71.01 &62.90 &73.06 & &92.46 &86.03 &77.26 &86.62 \\
& & & & & & & & & & & & \\
XLM-R-L &x,y &24 & &78.55 &74.83 &65.72 &74.46 & &90.15 &85.58 &85.98 &88.10 \\
XLM-R-L &x,y,x-y &24 & &75.22 &\textbf{75.80} &\textbf{69.01} &\textbf{74.66} & &90.27 &85.40 &85.50 &87.94 \\
XLM-R-L &x,y,x*y &24 & &80.55 &67.54 &63.38 &73.08 & &87.66 &81.48 &79.72 &84.08 \\
XLM-R-L &x,y,x*y,x-y &24 & &76.63 &73.76 &64.52 &72.65 & &91.26 &86.96 &\textbf{89.06} &\textbf{89.79} \\
XLM-R-L &SelfAttentive &24 & &73.17 &71.93 &62.14 &69.99 & &88.64 &\textbf{87.81} &80.86 &86.73 \\
XLM-R-L &MaxPooling &24 & &75.39 &72.33 &66.15 &72.26 & &89.30 &85.47 &85.39 &87.55 \\
\bottomrule
\end{tabular}
\caption{Experiment results for different multilingual pretrained models, in macro F1 score. We use bold font to highlight the maximum score across all settings and underline to highlight the maximum score in each part.}
\label{tab:all-results}
\end{table*}



\end{document}